%% file: main.tex
\documentclass[sigconf]{acmart}

\usepackage{float}
\usepackage{cleveref}
\usepackage{stfloats}
\usepackage{multirow}
\usepackage{array,graphicx}
\usepackage{booktabs}
\usepackage{pifont}

\usepackage{multicol}

\AtBeginDocument{%
  \providecommand\BibTeX{{%
    \normalfont B\kern-0.5em{\scshape i\kern-0.25em b}\kern-0.8em\TeX}}}

\begin{document}

\title{Fine-Tuning Large Language Models for Relation Extraction within a Retrieval-Augmented Generation Framework}




\author{Sefika Efeoglu}
\affiliation{%
  \institution{Freie Universitaet Berlin}
  \streetaddress{1 Th{\o}rv{\"a}ld Circle}
  \city{Berlin}
  \country{Germany}}
\email{sefika.efeoglu@fu-berlin.de}

\author{Adrian Paschke}
\affiliation{%
  \institution{Freie Universitaet Berlin}
  \city{Berlin}
  \country{Germany}\\
  \email{adrian.paschke@fokus.fraunhofer.de}
  \institution{Data Analytic Center (DANA)\\ Fraunhofer Institute FOKUS}
  \city{Berlin}
  \country{Germany}}








\input{sections/abstract}

\maketitle
\input{sections/introduction}
\input{sections/related_works}

\input{sections/methodology}

\input{sections/evaluation}

\input{sections/discussion}

\input{sections/conclusion}
\input{sections/limitations}
\onecolumn \begin{multicols}{2}
\bibliographystyle{ACM-Reference-Format}
\bibliography{sample-base}
\end{multicols}

\end{document}

%% file: sections/abstract.tex
\begin{abstract}
Information Extraction (IE) plays a pivotal role in transforming unstructured data into structured formats, such as Knowledge Graphs. One of the main tasks within IE is Relation Extraction (RE), which identifies relations between entities in text data. This process enriches the semantic understanding of documents, enabling more precise information retrieval and query answering. Recent works leveraging pre-trained language models have demonstrated significant performance improvements in RE. In the current era of Large Language Models (LLMs), fine-tuning these LLMs can mitigate the limitations of zero-shot RE methods, particularly in overcoming the domain adaptation challenges inherent in RE. This work explores not only the effectiveness of fine-tuned LLMs but also their integration into a Retrieval-Augmented Generation (RAG)-based RE approach to address domain adaptation challenges when general-purpose LLMs serve as generators within the RAG framework. Empirical evaluations on the TACRED, TACRED-Revisited (TACREV), and Re-TACRED datasets reveal substantial performance improvements with fine-tuned LLMs, such as Llama2-7B, Mistral-7B, and Flan-T5 Large and surpass previous methods on these datasets.

\end{abstract}
\keywords{
Relation Extraction,
Language Models,
RAG,
Information Extraction
}

%% file: sections/introduction.tex
\section{Introduction}
\label{sec:intro}
\begin{figure}[H]
    \centering
    \includegraphics[width=0.5\textwidth]{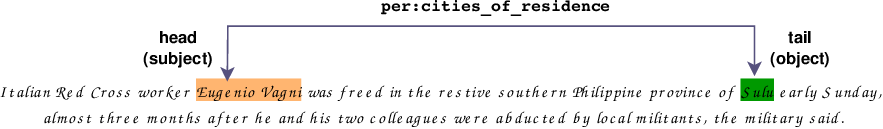}
    \caption{Representation of a relation, per:cities\_of\_residence, between head and tail entities in a sentence from the TACRED dataset.}
    \label{fig:into_tacred_examples}
\end{figure}

Information Extraction (IE) converts unstructured data into structured formats, such as Knowledge Graphs (KGs). A key IE task is Relation Extraction (RE), which identifies relationships between entities in text at the sentence level (See~\Cref{fig:into_tacred_examples}) or document level~\cite {Grishman_2015}. RE methods include supervised, unsupervised, and rule-based approaches~\cite{Aydar_2020,Sachin_2017}. Supervised RE methods generally yield strong performance but require extensive labeled data. However, recent studies show that RE methods using pre-trained language models (PLMs) can surpass traditional supervised approaches~\cite{zhou-chen-2022-improved,li2022reviewing,wang-etal-2022-deepstruct}. In the era of Large Language Models (LLMs), Retrieval-Augmented Generation (RAG)~\cite{gao2023retrieval,rag_meta} using zero-shot prompting settings, in-context learning~\cite{pan2024unifying}, or simple vanilla prompting~\cite{Zhang2023LLM-QA4RE}, have been utilized for RE without the need for additional model training.
The RAG-based prompting approach performs well when entity relations are easily derived from sentence tokens but struggles when relation types are not introduced into LLMs~\cite{efeoglu2024retrievalaugmented}. General-purpose LLMs, like Mistral~\cite{jiang2023mistral}, Llama2~\cite{Touvron2023Llama2O}, and Flan-T5~\cite{Flan_T5}, also show shortcomings in RE tasks due to insufficient domain-specific relation knowledge~\cite{efeoglu2024retrievalaugmented,Zhang2023LLM-QA4RE,xiong-etal-2023-rationale}. Incorporating these relation types into LLMs could enhance RE through zero-shot prompting~\cite{efeoglu2024retrievalaugmented}. To tackle this issue, we fine-tune language models on small sets of RE prompt datasets to enhance their ability to identify relations between entities at the sentence level. To evaluate the performance of fine-tuned LLMs, we conduct experiments using Llama2-7B~\footnote{https://huggingface.co/meta-llama/Llama-2-7b-chat-hf, accessed on 14.05.2025}~\cite{Touvron2023Llama2O},  Mistral-7B-Instruct-v0.2~\footnote{https://huggingface.co/mistralai/Mistral-7B-Instruct-v0.2, accessed on 14.05.2025}~\cite{jiang2023mistral}, and Flan T5 Large~\cite{Flan_T5} across three RE benchmark datasets: TACRED~\cite{zhang-etal-2017-position}, TACRED-Revisited (TACREV)\cite{alt-etal-2020-tacred} and Re-TACRED~\cite{Stoica_Platanios_Poczos_2021}. In this work, fine-tuning is used to overcome the limitations of zero-shot LLM prompting settings, such as RAG4RE~\cite{efeoglu2024retrievalaugmented}, in identifying relations between entities across TACRED and its variants. The contributions of our approach are as follows:

\begin{itemize} 
    \item Fine-tuning greatly improved LLM performance, with Flan-T5 Large outperforming larger models like Mistral-7B-Instruct-v0.2 and Llama2-7B on TACRED and its variants.
    \item Our fine-tuned LLMs, evaluated within RAG4RE, showed strong results on these datasets.
    \item This study is the first to fine-tune LLMs for the RE task and to systematically compare smaller and larger models like Mistral-7B-Instruct-v0.2 and Llama2-7B by parameter count.
\end{itemize}

The rest of this paper first summarizes RE approaches using the language models in~\Cref{sec:related_works} and then introduces our proposed approach~\footnote{The source code: \url{https://github.com/sefeoglu/fine-tuned-llm-relation-extraction}} in~\Cref{sec:methodology}. Afterwards, the experimental setup and results are presented in~\Cref{sec:evaluation} and  discussed in~\Cref{sec:discussion}. Lastly, all concluding remarks and future works are summarized in~\Cref{sec:conclusion}.

%% file: sections/related_works.tex
\section{Related Works}
\label{sec:related_works}
Relation Extraction (RE), as a core task of Information Extraction (IE), plays a significant role in natural language processing. RE aims to identify or classify the relations between (head and tail) entities in a given text. In this work, we primarily focus on sentence-level RE approaches.

RE can be achieved through various methods: supervised, unsupervised, distant supervision, weak supervision, and rule-based~\cite{Sachin_2017}. Supervised methods require costly, annotated data~\cite{Sachin_2017}; distant supervision reduces data needs but risks noise~\cite{Aydar_2020}; weak supervision may lead to semantic drift~\cite{Eugene_2000}; and rule-based methods are limited by predefined rules~\cite{Sachin_2017}. In addition to the fundamental approaches, leading RE methods with fine-tuned LLMs include Cohen et al.'s span prediction for broader entity relations~\cite{cohen2020relation}, DeepStruct's structural enhancements, Zhou et al.'s entity-aware self-attention~\cite{zhou-chen-2022-improved}, and Li et al.'s label graph for top-K prediction analysis~\cite{li2022reviewing}. Furthermore, Zhang et al.~\cite{Zhang2023LLM-QA4RE} used multiple-choice prompts, improving RE predictions with added context, though it does not surpass prior rule-based methods. Chen et al.~\cite{CHEN2024123478} introduced context-aware prompt tuning, while RAG4RE~\cite{efeoglu2024retrievalaugmented} utilized retrieval-augmented prompting, all tested on TACRED and similar benchmarks.

In this work, we aim to fine-tune LLMs on RE prompt datasets to improve domain adaptation and evaluate their performance on benchmark datasets.

%% file: sections/methodology.tex
\section{Methodology}
\label{sec:methodology}

This work addresses the challenge of sentence-level RE using general-purpose LLMs with zero-shot prompting. General-purpose LLMs struggle with domain-specific relation types, so we fine-tune them on a small RE prompt dataset to improve their ability to identify entity relations. We detail the fine-tuning process in~\Cref{sec:met_fine-tuning} and describe the integration of fine-tuned LLMs into the RAG4RE approach in~\Cref{sec:met_rag_re}.
\begin{figure}[!htb]

   \begin{minipage}{0.5\textwidth}
     \centering
     \includegraphics[width=\linewidth]{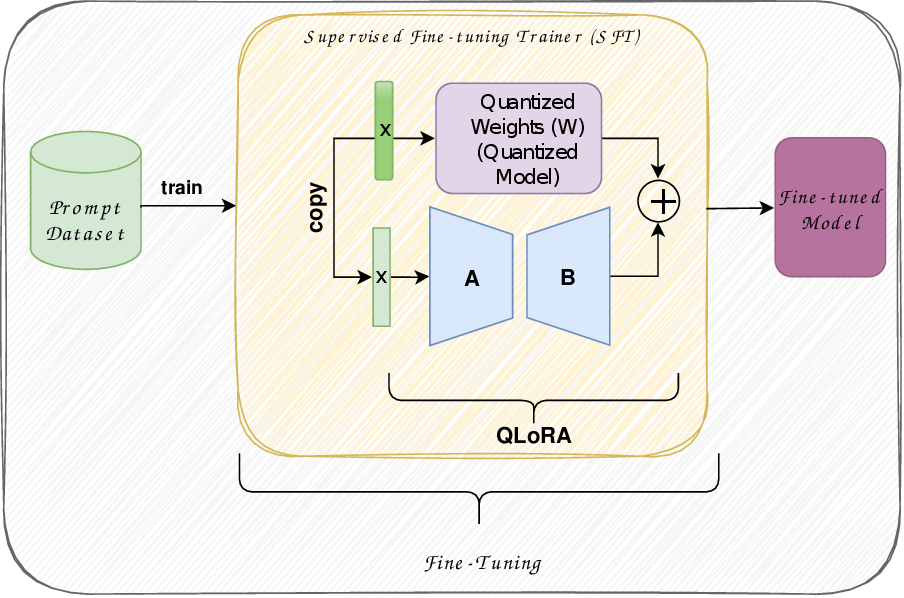}
     \caption{Fine-tuning a pre-trained model on a prompt dataset alongside the QLoRA adapter and SFT.}
     \label{fig:fine-tune}
   \end{minipage}\hfill
   \begin{minipage}{0.5\textwidth}
     \centering
     \includegraphics[width=\linewidth]{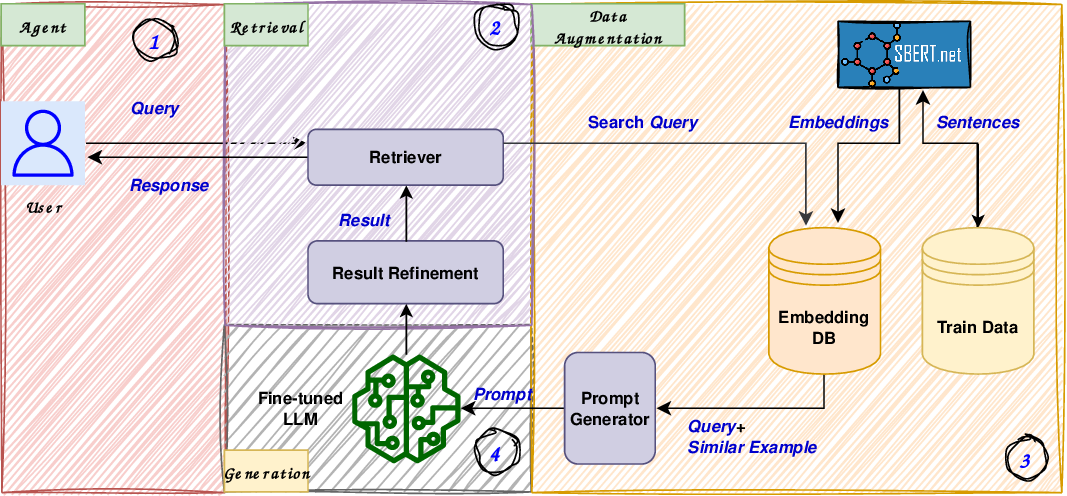}
     \caption{RAG with fine-tuned Large Language Models.}
     \label{fig:ragwithmodel}
   \end{minipage}

\end{figure}

\subsection{Fine-tuning Models on Prompt Datasets}
\label{sec:met_fine-tuning}
We fine-tune both encoder-decoder models (such as Flan-T5) and decoder-only models, e.g., Llama2-7B and Mistral-7B, on RE prompt datasets using the Supervised Fine-Tuning Trainer (SFT)~\footnote{SFT:~\url{https://huggingface.co/docs/trl/sft\_trainer}}. This fine-tuning process facilitates domain adaptation for general-purpose LLMs. The SFT approach, which requires labeled training data, is straightforward to implement and train. Additionally, we utilize the Low-Rank Adaptation for quantized language models (QLoRA) method~\cite{dettmers2023qlora} to fine-tune LLMs. QLoRA optimizes model parameters for text generation while minimizing memory usage on GPUs, which is crucial in scenarios with limited GPU memory. 
\noindent
\subsubsection{Prompt Dataset Generation.}
\begin{figure}[H]
    \centering
    \includegraphics[width=0.5\textwidth]{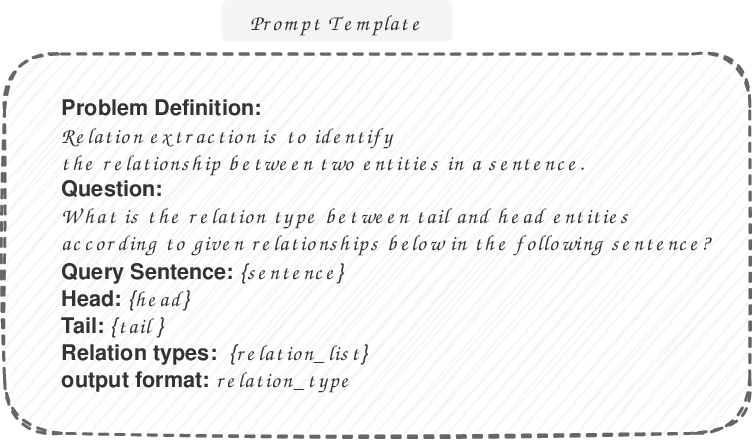}
    \caption{A prompt template for fine-tuning a Large Language Model.}
    \label{fig:prompt_dataset_template}
\end{figure}

The RE prompt dataset is constructed following the template outlined in a previous study by~\cite{efeoglu2024retrievalaugmented}. This dataset originates from a supervised dataset within a single domain and utilizes a specialized template for fine-tuning, as illustrated in~\Cref{fig:prompt_dataset_template}.

\noindent
\subsubsection{Parameter Efficient Fine-Tuning.}
We utilize QLoRA, a parameter-efficient fine-tuning method that begins by applying quantization to a pre-trained language model. This technique reduces the model's high-precision floating-point representation to a lower precision, thus decreasing memory usage. In particular, we use the ``4-bit NormalFloat (NF4)'' format, which is optimized for normally distributed data and has been shown to outperform traditional 4-bit integers and floats~\cite{dettmers2023qlora}. Following quantization, LoRA is applied to specific model modules. Fine-tuning is subsequently conducted using the SFT on a single-domain, task-specific prompt dataset. The entire process is illustrated in~\Cref{fig:fine-tune}.

\subsection{ Retrieval-Augmented Generation with Fine-Tuned Models}
\label{sec:met_rag_re}
The Retrieval-Augmented Generation-based Relation Extraction (RAG4RE) approach, introduced by~\cite{efeoglu2024retrievalaugmented}, comprises three modules: i.) Retrieval, ii.) Data Augmentation, and iii.) Generation. In our implementation, we integrate fine-tuned LLMs, trained on RE prompt datasets, into the generation module of the RAG4RE approach~\cite{efeoglu2024retrievalaugmented} to address the task of identifying relations between entities in sentences, as illustrated in~\Cref{fig:ragwithmodel}. Specifically, the LLM used in the generation module of RAG4RE is replaced with our fine-tuned LLMs, while all other components of RAG4RE remain unchanged.

%% file: sections/evaluation.tex
\section{Evaluation}
\label{sec:evaluation}

We evaluate our approach using three benchmark datasets and language models. In~\Cref{sec:experimental_setup}, we detail the datasets, metrics, and experimental settings, including the fine-tuning of language models and the use of Retrieval-Augmented Generation with these fine-tuned models. Then, we present and analyze the experimental results, comparing them with those of previous high-performing RE methods in~\Cref{sec:results}.

\subsection{Experimental Setup}
\label{sec:experimental_setup}
In this section, we initially introduce the datasets utilized for evaluation, followed by the detailed settings used for fine-tuning and the RAG4RE framework~\cite{efeoglu2024retrievalaugmented}, leveraging our fine-tuned language model within its generation module.
\newline
\noindent
\textbf{Datasets.} We utilize three RE benchmark datasets: TACRED~\cite{zhang-etal-2017-position}, TACREV~\cite{alt-etal-2020-tacred}, and Re-TACRED~\cite{Stoica_Platanios_Poczos_2021} as detailed in~\Cref{tab:dataset}. The prompt datasets are generated from the validation partitions of the benchmark datasets. The training datasets are utilized in the Embedding Database (DB) of RAG4RE~\cite{efeoglu2024retrievalaugmented}, while the test splits are used for evaluation. We ensure a strict separation between the training and test splits across all benchmark datasets.
\begin{table}[H]
\tiny
    \centering

    \caption{The table gives the number of sentences in the test, train, and prompt datasets, as well as the number of relations per benchmark dataset.}
    \label{tab:dataset}
    \begin{tabular}{p{3cm}| c c c}
    \toprule
    \textbf{Split}&\textbf{TACRED}&\textbf{TACREV}&\textbf{Re-TACRED}\\
    \midrule
    Train&68124&68124&58465\\
    Test&15509&15509&13418\\
    Validation&22631&22631&19584\\
    Prompt Dataset (Generated from Validation)&22631&22631&19584\\
    \textit{\#} of Relations &42&42&40\\
    \bottomrule
    \end{tabular}

    \end{table}
\noindent
\subsubsection{Metrics}
\label{sec:metrics}
The benchmark datasets used—TACRED and its variants—are imbalanced, with a high proportion of ``no\_relation'' labels~\cite{alt-etal-2020-tacred,Stoica_Platanios_Poczos_2021}, necessitating the use of micro metrics. For instance, in the TACRED test split, 12,184 out of 15,509 relations are labeled as ``no\_relation''. We evaluate our experiments using the micro F1-score, precision, and recall across all three benchmark datasets.
\newline
\subsubsection{Settings for Models}
\label{sec:setting_fine_tuning}  
We employed the fine-tuning approach from~\Cref{sec:met_fine-tuning}, using a single GPU with 48 GB of memory and the parameters detailed below. Building on prior studies in RE with language models~\cite{efeoglu2024retrievalaugmented,Zhang2023LLM-QA4RE}, we utilized the following LLMs:  
\begin{itemize}  
    \item [--]Flan T5 Large: An encoder-decoder model~\cite{Flan_T5,pan2024unifying} with 770M parameters. LoRA parameters: alpha=32, dropout=0.01, r=4. Hyperparameters: learning rate=5e-5, batch size=8, one epoch.  
    \item [--] Mistral-7B~\cite{jiang2023mistral,pan2024unifying} and Llama2-7B~\cite{pan2024unifying,Touvron2023Llama2O}: Decoder-only models with 7B parameters, used in~\cite{efeoglu2024retrievalaugmented}. We used Mistral-7B-Instruct-v0.2~\footnote{https://huggingface.co/mistralai/Mistral-7B-Instruct-v0.2}. LoRA parameters: alpha=16, dropout=0.1, r=64. Hyperparameters: learning rate=2e-4, batch size=4, one epoch, weight decay=0.001.  
\end{itemize}

\noindent
\subsubsection{Settings for RAG4RE}
\label{sec:setting_rag_re}
Due to limited GPU resources, we were unable to fine-tune the Flan-T5 XL model used in the original RAG4RE~\cite{efeoglu2024retrievalaugmented}. Therefore, all experimental settings are replicated from RAG4RE with Flan-T5 Large. We strictly adhere to the experimental setups established in RAG4RE for our study.
\input{sections/results}

%% file: sections/results.tex
\subsection{Results }
\label{sec:results}

We evaluated language models fine-tuned on prompt datasets detailed at~\Cref{tab:dataset} in~\Cref{sec:experimental_setup}. Furthermore, we integrated these fine-tuned language models into the RAG4RE~\cite{efeoglu2024retrievalaugmented}. It is worth noting that due to constraints in GPU resources, we opted to utilize Flan-T5 Large instead of Flan-T5 XL or XXL for fine-tuning. Hence, we chose Flan-T5 Large and meticulously replicated the RAG4RE experiments within the confines of our work. In this section, we first introduce the results of our fine-tuned models and then the results of RAG4RE approach using our fine-tuned models.

With regard to the evaluation of fine-tuned LLMs alongside LoRA on three different datasets, fine-tuned Mistral-7B models accomplish outstanding performance at~\Cref{tab:sota_results}. Notably, these fine-tuned Mistral-7B models achieve remarkable F1 scores of 89.64\%, 94.61\%, and 90.09\% on TACRED, TACREV, and Re-TACRED, respectively (see~\Cref{tab:sota_results}). The Llama2-7B models fine-tuned on TACRED and TACREV follow the fine-tuned Mistral-7B models with micro-F1 scores of 88.20\% and 93.75\%. Unfortunately, the fine-tuned Llama2-7B models could not exhibit the same performance on Re-TACRED at~\Cref{tab:sota_results}. The fine-tuned Flan-T5 Large model takes second place with an F1 score of 86.94\% on the Re-TACRED dataset (see~\Cref{tab:sota_results}). Moreover, fine-tuning LLMs outperformed simple query prompting and the previously introduced RAG4RE method~\cite{efeoglu2024retrievalaugmented}.
Additionally, we integrated these fine-tuned LLMs into the RAG4RE approach~\cite{efeoglu2024retrievalaugmented} in order to explore their potential in addressing the limitations of general-purpose LLMs. 

Remarkably, the integration of fine-tuned models into RAG4RE yielded significant improvements across all three datasets, including TACRED, TACREV and Re-TACRED, particularly when leveraging Flan-T5 Large at~\Cref{tab:sota_results}. While we observed enhancements in RAG4RE's performance, as detailed in~\cite{efeoglu2024retrievalaugmented}, with the integration of fine-tuned Llama-7B on Re-TACRED, it is noteworthy that this improvement was not observed on TACRED and TACREV. Regrettably, the results indicate that the use of Mistral-7B as the fine-tuned LLM did not yield improvements in the results of RE. The reason why the performance of the RAG4RE approach could not be improved when fine-tuned decoder-only models are used as a generator in its architecture (see~\Cref{fig:ragwithmodel}) might be related to catastrophic forgetting. Previous work fine-tuning language models on a single task is also dealing with the same forgetting problem~\cite{feng2024mixtureofloras}.

As a result, the fine-tuned Flan-T5 Large models consistently achieved the highest F1 scores among all the experiments conducted in this work, particularly when integrated into the RAG4RE framework proposed in~\cite{efeoglu2024retrievalaugmented}. However, fine-tuned Mistral is slightly better than RAG4RE using fine-tuned Flan-T5 Large on TACREV. In addition to the findings of the experiments using Flan-T5 Large, both fine-tuning language models on the dataset and integrating these fine-tuned models into RAG4RE outperformed zero-shot prompting approaches, such as simple queries and RAG4RE~\cite{efeoglu2024retrievalaugmented} (see~\Cref{tab:sota_results}).
\begin{table*}[htbp]
\tiny
  \centering
  \caption{Experimental results on three benchmark datasets using different large language models (LLMs) and methods.}
  \label{tab:sota_results}
  \begin{tabular}{p{1.5cm}|l|rrr|rrr|rrr}
    \toprule
    \multicolumn{2}{c|}{\textbf{}} & \multicolumn{3}{c|}{\textbf{TACRED}} & \multicolumn{3}{c|}{\textbf{TACREV}} & \multicolumn{3}{c}{\textbf{Re-TACRED}} \\
    \cmidrule(lr){3-5} \cmidrule(lr){6-8} \cmidrule(lr){9-11}
    \textbf{LLM} & \textbf{Method} & \textbf{P (\%)} & \textbf{R (\%)} & \textbf{F1 (\%)} & \textbf{P (\%)} & \textbf{R (\%)} & \textbf{F1 (\%)} & \textbf{P (\%)} & \textbf{R (\%)} & \textbf{F1 (\%)} \\
    \midrule
    \textit{T5 Large} 
    & Simple Query & 95.10 & 03.18 & 06.16 & 96.72 & 06.90 & 12.89 & 90.91 & 00.26 & 00.51 \\
    & RAG4RE & 85.99 & 34.50 & 49.20 & 91.28 & 08.20 & 15.04 & 80.77 & 00.27 & 00.53 \\
    & Fine-tuning (QLoRA) & 86.74 & 86.76 & 86.74 & 89.93 & 90.13 & 90.03 & 86.27 & 87.62 & 86.94 \\
    & RAG4RE + Fine-tuning & 89.93 & 94.17 & \textbf{92.00} & 95.02 & 93.66 & 94.34 & 92.31 & 93.73 & \textbf{93.01} \\
    \hline
    \textit{LLaMA2-7B} 
    & Simple Query~\cite{efeoglu2024retrievalaugmented} & 84.97 & 01.21 & 02.38 & 74.64 & 00.44 & 00.87 & 80.20 & 00.94 & 01.86 \\
    & RAG4RE~\cite{efeoglu2024retrievalaugmented} & 81.23 & 55.01 & 65.59 & 84.89 & 54.57 & 66.43 & 55.93 & 03.46 & 06.52 \\
    & Fine-tuning (QLoRA) & 88.07 & 88.34 & 88.20 & 90.07 & 97.73 & 93.75 & 87.54 & 44.58 & 59.08 \\
    & RAG4RE + Fine-tuning & 80.29 & 89.18 & 84.50 & 84.10 & 97.26 & 90.22 & 83.53 & 68.16 & 75.07 \\
    \hline
    \textit{Mistral-7B} 
    & Simple Query~\cite{efeoglu2024retrievalaugmented} & 94.67 & 11.96 & 21.23 & 92.34 & 05.15 & 09.75 & 64.64 & 05.48 & 10.11 \\
    & RAG4RE~\cite{efeoglu2024retrievalaugmented} & 87.81 & 30.10 & 44.83 & 93.23 & 22.59 & 36.36 & 60.19 & 30.08 & 40.11 \\
    & Fine-tuning (QLoRA) & 94.73 & 85.06 & 89.64 & 95.79 & 93.48 & \textbf{94.61} & 92.40 & 87.83 & 90.09 \\
    & RAG4RE + Fine-tuning & 86.57 & 82.88 & 84.68 & 97.58 & 79.33 & 87.50 & 90.86 & 85.95 & 88.33 \\
    \bottomrule
  \end{tabular}
\end{table*}

%% file: sections/discussion.tex
\section{Discussion}
\label{sec:discussion}
Our findings demonstrate significant improvements over the original RAG4RE~\cite{efeoglu2024retrievalaugmented} results on the TACRED, TACREV, and Re-TACRED datasets, as shown in~\Cref{tab:sota}, when fine-tuned Flan-T5 Large models are integrated into the RAG4RE approach. Fine-tuning language models, particularly in the context of domain adaptation, led to substantial performance enhancements for both general-purpose LLMs and RAG4RE~\cite{efeoglu2024retrievalaugmented} (see~\Cref{tab:sota}). The F1 scores of RAG4RE combined with fine-tuned LLMs surpassed those of previous approaches across all three datasets, as illustrated in~\Cref{tab:sota}. Similarly, the F1 scores of the fine-tuned LLMs exceeded those of prior approaches that employed both zero-shot prompting and pre-trained language models (PLMs) (see~\Cref{tab:sota}). The best-performing results in our experiments, reported in~\Cref{tab:sota}, surpassed those of approaches using both zero-shot prompting and PLMs on the TACRED, TACREV, and Re-TACRED datasets, achieving F1 scores of 92.00\%, 94.61\%, and 93.01\%, respectively. Furthermore, our RAG4RE+Fine-tuning approach also outperformed the original RAG4RE utilizing general-purpose LLMs. Therefore, our fine-tuned LLMs achieved outstanding results on the TACRED, TACREV, and Re-TACRED datasets when integrated into the RAG4RE framework~\cite{efeoglu2024retrievalaugmented}.
\begin{table*}[htbp]
\tiny
  \centering
  \caption{A comparison of our best-performing results with those of prior works in terms of F1-score.}
  \label{tab:sota}
  \scriptsize
  
  \begin{tabular}{llccc}
    \toprule
  \textbf{Method Type}&\textbf{Method}& \textbf{TACRED}&\textbf{TACREV}&\textbf{Re-TACRED}\\
    \cmidrule{1-5}
     &DeepStruct~\cite{wang-etal-2022-deepstruct}&$76.8\%$&-&-\\
     &EXOBRAIN~\cite{zhou-chen-2022-improved} &$75.0\%$&-&91.4\%\\
\textbf{PLM-based}  &KLG~\cite{li2022reviewing}&- &$84.1\%$&-\\
     &SP~\cite{cohen2020relation} &$74.8\%$ &-& \\
      &GAP~\cite{CHEN2024123478} &$72.7\%$ &82.7\%& 91.4\% \\
    
     \cmidrule{1-5}
         &LLMQA4RE~\cite{Zhang2023LLM-QA4RE}&$52.2\%$&$53.4\%$&$66.5\%$\\
    \textbf{Zero-Shot prompting}&RationaleCL~\cite{xiong-etal-2023-rationale}&$80.8\%$&-&-\\
    &RAG4RE~\cite{efeoglu2024retrievalaugmented}&$86.6\%$&$88.3\%$&$73.3\%$\\
    \cmidrule{2-5}
    \cmidrule{2-5}
   & \textbf{RAG4RE+Fine-tuning (Ours)}&\textbf{92.00\%}&$\underline{94.34}\%$&\textbf{93.01\%}
    \\
    \cmidrule{2-5}
    & \textbf{Fine-tuning (Ours)}&$\underline{89.64}\%$&\textbf{94.61\%}&$90.09\%$\\
    \bottomrule
  \end{tabular}

 \end{table*}

%% file: sections/conclusion.tex
\section{Conclusion}
\label{sec:conclusion}

We address domain adaptation challenges in zero-shot relation extraction (RE) with general-purpose LLMs by fine-tuning Flan-T5 Large, Mistral-7B-Instruct-v0.2, and Llama2-7B on TACRED, TACREV, and Re-TACRED datasets. Our fine-tuned models outperformed previous methods, including RAG4RE~\cite{efeoglu2024retrievalaugmented}. Integrating these fine-tuned LLMs into RAG4RE significantly enhanced its performance, especially with Flan-T5 Large. However, Llama2-7B and Mistral-7B showed inconsistent F1 scores, likely due to single-task fine-tuning issues. Future work will explore multi-task fine-tuning for RE and entity recognition to mitigate catastrophic forgetting~\cite{feng2024mixtureofloras,liu2023mftcoder,yang2024unveiling}.

%% file: sections/limitations.tex
\section*{Limitations}
\label{sec:limitations}
This approach requires an embedding database within the data augmentation module of the RAG and retrieves the most similar sentence for use in the RAG module. The most similar sentence with the sentence in the query might have low similarity score. The pre-trained language models may already be familiar with these datasets, as noted in~\cite{efeoglu2024retrievalaugmented}, since they might be trained on these benchmark datasets.